# Alpha-Net: Architecture, Models, and Applications

**Adya Sharma**  
MANIT, Bhopal

**Jishan Shaikh**  
MANIT, Bhopal

**Ankit Chouhan**  
MANIT, Bhopal

**Avinash Mahawar**  
MANIT, Bhopal

## Abstract

Deep learning network training is usually computationally expensive and intuitively complex. We present a novel network architecture for custom training and weight evaluations. We reformulate the layers as ResNet-similar blocks with certain inputs and outputs of their own, the blocks (called Alpha blocks) on their connection configuration form their own network, combined with our novel loss function and normalization function form the complete Alpha-Net architecture. We provided empirical mathematical formulation of network loss function for more understanding of accuracy estimation and further optimizations. We implemented Alpha-Net with 4 different layer configurations to express the architecture behavior comprehensively. On a custom dataset based on ImageNet benchmark we evaluate Alpha-Net v1, v2, v3, and v4 for image recognition to give accuracy of 78.2%, 79.1%, 79.5%, and 78.3% respectively. The Alpha-Net v3 gives an improved accuracy of approx. 3% over last state-of-the-art network ResNet 50 on ImageNet benchmark. We also present analysis on our dataset with 256, 512, and 1024 layers and different versions of the loss function. The input representation is also very crucial for training as initial preprocessing will take only a handful of features to make training less complex than it needed to be. We also compared network behaviour with different layer structures, different loss functions, and different normalization functions for better quantitative modeling of Alpha-Net.

## 1 Introduction and Motivation

Deep learning is a division of machine learning in artificial intelligence (AI) that mimics the structure of the human brain in processing data and generating patterns to use in making important decisions. It is an artificial intelligence operation that has a meshwork of neurons capable of learning associations for unseen from data that is structured or unstructured. It is hence rightly known as deep neural learning or deep neural network.

In Deep Convolutional Neural Networks (CNN), the introductory layers axiomatically learn the features laid out for years or even decades, and the subsequent layers further learn advanced level abstraction. Ultimately, the combination of these advanced level abstractions depicts facial identity with unprecedented stability.

The principal distinguishing element of deep learning compared to more conventional approaches is the ability of the effectiveness of the classifiers to go large scale with an increase in volume of data. Former machine learning methods typically plateau in performance after it reaches a threshold of training data. Deep learning is a distinctive algorithm whose performance enhances with increase in data as more the data fed, the better the classifier is trained on, resulting in surpassing the classical models/algorithms.

## 1.1 Problem Definition

Deep learning network training is a tedious task because of high computational resources and large number of functions to be tuned. Training purposes mainly for face recognition and object detection methods. As yet, various methods have been designed for face recognition, but it still has remained extremely challenging in real life applications and till date, there is no technique which parallels human ability to distinguish faces despite countless diversity in appearance that the face can take in a scene and implement a powerful solution to all situations accurately irrespective of variations due to light, aging, expressions, similarity in faces and other methodical problems like noise, image acquisition, video camera distortion. Each and every method invented till date suffers from limitations, which hinder the process of achieving cent percent accuracy. **The main task of the project is create a novel DL based architecture, with which we can create separate models for use-cases and use it for our applications.**

Older machine learning algorithms typically plateau in performance after it reaches a threshold of training data. In this scenario, deep learning algorithms come to the rescue as they provide high performance even with an increase in data. In fact, it follows two principles: "The more the data, the better the model" and the "deeper the network, the more powerful it gets".

**In these kind of situations, it is imperative to enhance the training methods by introducing a new method using more powerful and efficient functions and architectures, which are robust enough to work for all kinds of variations, yielding better accuracy in less time.**

## 1.2 Research Objectives

In order to overcome the drawbacks of the already existing algorithms, we based our method of creating a new architecture that can train complex networks efficiently and in optimal manner.

The method is based on an improvement over the original ResNet architecture, which has been a benchmark architecture for computer vision tasks such as object detection, face recognition, and segmentation.

The novelties in the project can be summarized as:

1. GAN based dataset for more data; improves training accuracy.
2. Alpha-encodings and transformations of dataset items; for near optimal dataset size and features.
3. New method of interconnection between blocks based on stochastic (random) nature from data items; for better training results in slower time
4. Combined results of convolution filters (e.g. average of 5*5 and 10*10 filters) at each block of the network;
5. Loss function (optimizer); A new variant of AM Softmax function.

The objective of this work is to define a technique robust to the noise to represent, detect and recognize human faces efficiently. It is shown that the described strategy supports not only to acquire and perceive the examples provided during the training phase, but also to generalize them, thus enabling us to detect occurrences of the activities that have not been enclosed in the training set. The expected results show that our system reliably recognizes sequences of human faces with a high recognition rate.

## 2 Related work

### 2.1 Residual Networks (ResNet)

Residual Networks are one of the most studied networks in neural network architectures. The basic idea of ResNet is to train large number of layers of a neural network efficiently and in a lesser time. ResNet gives an effective analysis of why increasing number of layers at certain point decreases accuracy. The compelling performance of ResNet and its variants can be explained by its residual blocks and identity mappings. The powerful and complex representational ability of ResNet allows computer vision tasks efficiently and easily such as object detection, face recognition, and image segmentation. Figure 1 shows the decrease in accuracy for 20-layer and 56-layer ResNet architecture.

Universal approximation theorem states that any single layer of feed forward network with sufficient capacity can represent any function. However, the layer will get enormous and the network becomes prone to overfitting the data. The trend in community blasts of increasing layers increases training comprehensively, but increasing layers did decrease the training error at higher number of iterations.

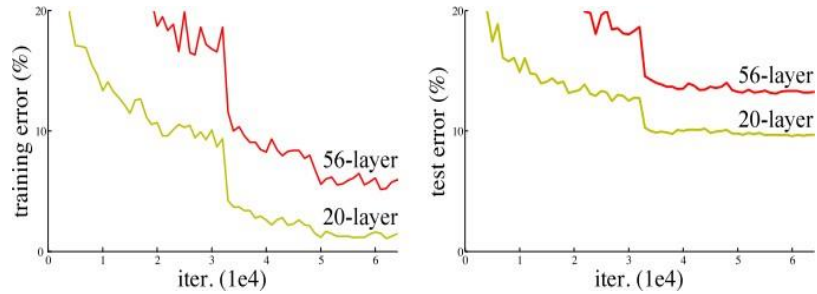

Figure 1: Increasing layers decrease accuracy as shown by iterative graphs of training error vs number of iterations performed; 56-layer performed worse than 20-layer network for ResNet architecture according to K. He et al.

Vanishing gradient problem does not let Deep NNs to easily train in most cases – the problem that leads to huge trouble in training neural networks for any fundamental task. Result of which, the performance of it became saturated and even degraded in some of the tasks.

Early attempts to handle the vanishing gradient problem is to add an auxiliary loss layer in the network to adjust the gradient of the network at each iteration of training. The core idea of ResNet is to bring a block of residual network lapsed with different parts such as ReLU, etc. and activation layer, and connection of which leads to the complete network.

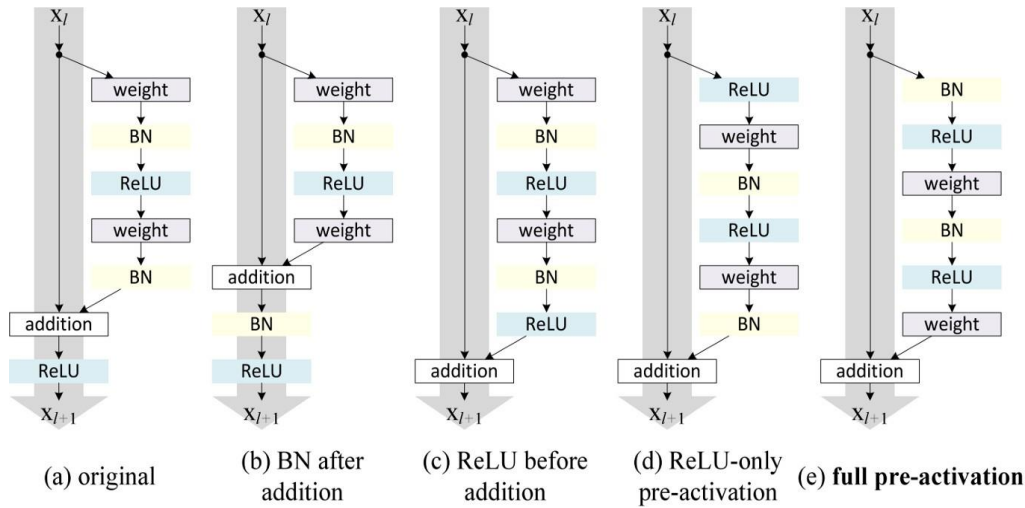

Figure 2: Different variants of ResNet proposed (a) original one with BN before addition, (b) BN after addition and after weight, (c) ReLU layer before addition and before weight, (d) Use of ReLU layer only for pre-activation, and (e) full pre-activation layers order. Figure by K. He et al.

The variants of ResNets can be explained by changing its block nature, size, and layers including the BN after addition, ReLU before addition, ReLU - only pre-activation, and full-activation.

## 2.2 ResNeXt

The authors introduced a concept called cardinality, to ensure a novel way of regulating the model capacity. They showed that accuracy can be gained more methodically by increasing the cardinality, instead of going deeper or wider network-wise. The authors testified that this network framework is relatively easy to train as compared to Inception Module of GoogleNet since it requires only one hyperparameter, and Inception module requires multiple hyperparameters to tune.

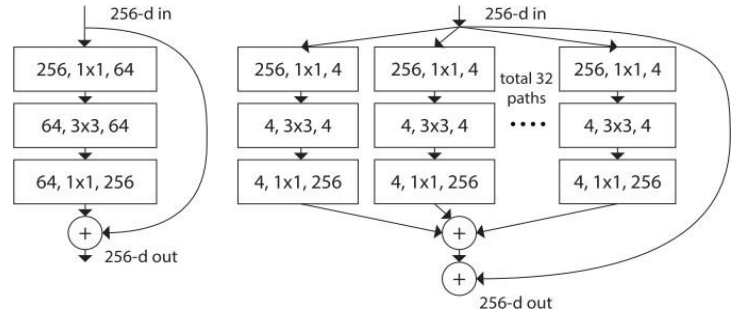

Figure 3: Block structure of ResNext as proposed by S. Xhang et al. The size of input vector is same as the size of output vector (256-dimensional) with (a) single path of data flow and (b) multiple (32) paths of the data flow. The output of which will be followed by addition layer.

The "split-transform-merge" is commonly done by pointwise grouped convolutional layer, which partitions its input into groups of feature maps and perform normal convolution respectively, then their outputs are depth-concatenated and then sent as input to a 1x1 convolutional layer.

## 2.3 Densely connected CNN

Also called as DenseNet, it connects all layers directly with each other. In this contemporary framework, the input of each layer is composed of the feature maps of all former layers, and its output is fed to each subsequent layer. The feature maps are then combined using depth-concatenation.

Apart from dealing with the vanishing gradient problem, the authors state that this architecture also promotes feature reuse, making the network extremely parameter-efficient. One straightforward analysis of this is that, the output of the identity mapping was supplemented to the next blocks, which might deter information flow if the feature maps of two layers have very different distributions.

## 2.4 Deep Network with Stochastic Depth

In spite of ResNet proving its effectiveness in many applications, its one prominent drawback is that deeper networks usually need weeks of training, rendering it practically infeasible in real life applications. To handle this problem, Huang et al. devised a counter-intuitive approach of randomly dropping layers during training, and using the entire framework in testing.

They used only residual blocks for network architecture, the mapping was not particularly identity, the flow can be 2-way in the network. **This inspires us to search for stochastic mapping with convenient network size.** The survival probability is randomly dropped during training time.

## 3 Methodology

### 3.1 Research Methodology

The most important steps of the proposed method can be summarized in the following points:

- Dataset organization: ImageNet + CASIA-webface + GAN based dataset + some random images.
- Face alignment using landmark algorithms.

- Input normalization using Alpha-transformations.
- Custom encoding of normalized Numpy matrices for blocks.
- Assigning random weights to interconnect layers initially.
- Add linear weights + Additive margin softmax as loss function layer
- Alpha-Net architecture for training data
- Results combination and analysis

### 3.2 Work Description

#### 3.2.1 Dataset

We are augmenting a dataset of human faces using GAN. The input to this step is the CASIA-WebFace dataset which is the largest dataset publicly available.

Discriminative algorithms tend to try to classify input data; that is, given the features of an occurrence of data, they try to conclude a category or label to which that data corresponds. For instance, given the set of words in an email (the data instance), a discriminative algorithm can conclude as to whether the message is spam or not. Therefore, discriminative algorithms map features to categories or labels. They are interested entirely with that correlation.

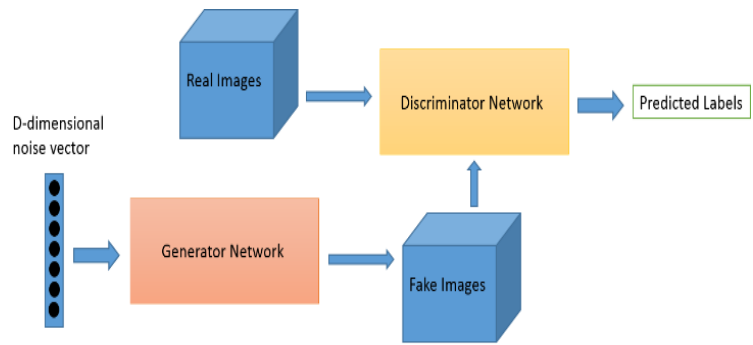

Figure 4: Data generation using Generative adversarial network. The connection between discriminator network and generator network is sequentially aligned like a linked list structure. The output of the system is given by discriminator network as predicted labels as expected. Ref: GAN (Facebook) paper.

Generative adversarial networks (GANs) are deep neural net framework composed of two networks, pitting one against the other (thus called "adversarial"). Rather than assigning a label given certain features, they try to conclude features given a certain label. Generative models design the distribution of separate classes. One neural net, named the generator, generates novel data occurrences, while the other, known as the discriminator, gauges them for legitimacy; i.e. the discriminator determines whether each occurrence of data that it assesses belongs to the actual training dataset or not.

The GAN dataset + CASIA webface dataset + ImageNet dataset is then combined with some random images for more diverseness, and combined dataset is then used for training.

#### 3.2.2 Face alignment

When we clip an image of a person or object, the direction of the person/object facing as well as the position of camera matters a lot. Sometimes other factors such as luminosity and image quality also affect the qualitative features of the image/face. The face orientation is a complex problem, but we can use a simple template of 68 landmark points to assign the face with each point for mapping purposes. After this we can wrap the picture with respect to the template only, and later on we can only use template image features for training and classification purposes.

When we map a face with a template we know where all the biometrics are. We can then use affine transformations (rotation, translation, and shear) for verification of the face alignment and mapping.

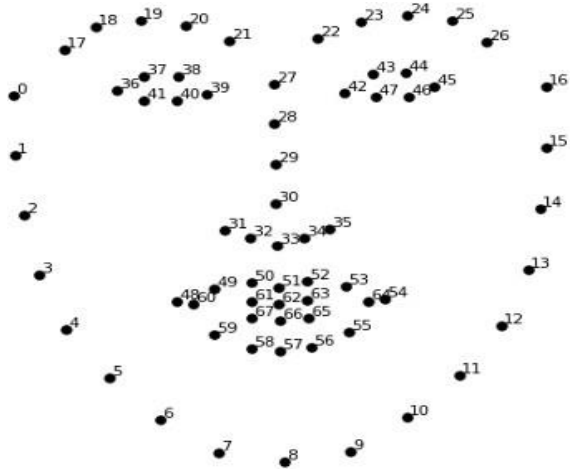

Figure 5: 68 landmark points for alignment. Too many landmark points will often lead to inclusive results and more true negative cases for verification. Ref: Openface Implementation.

Often the case, the image is usually 30 to 60 degree tilted either to right or left, which can be easily aligned by rotation or transformation used properly with size.

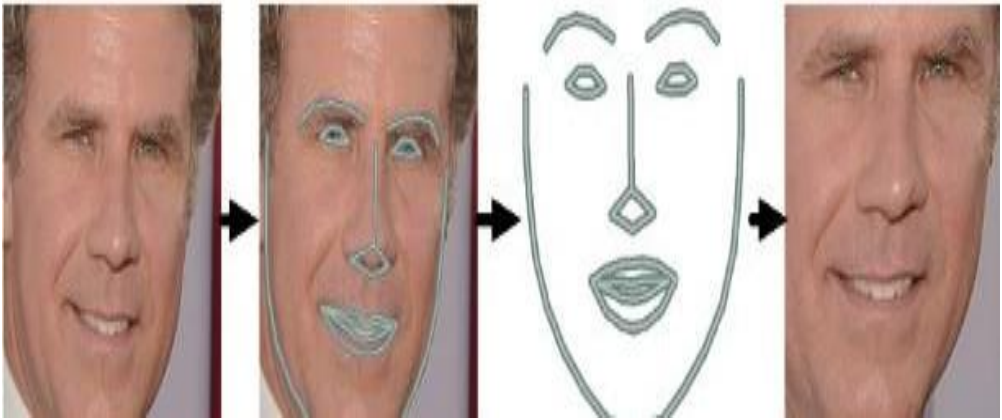

Figure 6: Face alignment on a sample image.

The orientation of the face in the data image does not matter, as we can map it to the given particular template. The same landmark points can be used for verification and authentication purposes at later stage, if used manually.

### 3.2.3 Loss function

Softmax function is widely used in computer vision and deep learing community for classification and learning tasks. This is because softmax function provides less inter-class correlation, and high intra-class correlation. We modified the Additive margin version of Softmax function, to add linear weights trained specially for ImageNet benchmark. The formulation of AM Softmax via $cos(\theta - m)$, can be mathematically noted as:

$$\Lambda_{AMS} = -\frac{1}{n}\sum_{i} \frac{e^{s(\cos\theta_{y_i} - m)}}{e^{s(\cos\theta_{y_i} - m)} + \sum_{c, j=1, j \neq y}^{} e^{s\cos\theta_j}}$$

For linear weights ($y = ax_i + c$), we can add a piece-wise division of it to make AM Softmax function with linear weights trained on ImageNet benchmark, as follows:

$$\Lambda_{AMS} = \begin{cases} -\frac{1}{n}\sum_{i=1}^{n} \frac{e^{s(\cos\theta_{y_i} - m)}}{e^{s(\cos\theta_{y_i} - m)} + c\sum_{j=1, j\neq y}e^{s\cos\theta_j}} & \theta - m > 0 \\ ax_i + c & \theta - m <= 0 \end{cases}$$

Where:

$\Lambda_{AMS}$ = Calculated Loss

$n$ = Training instances

$\theta$ = Angle with the origin

$m$ = Gradient for the instance

$s$ = Sample value of the current instance

$a$ and $c$ = Linear weight coefficients

### 3.2.4 Alpha-Net Architecture

We have designed 4 type alpha nets, and have observed comprehensive results for our dataset. The two basic types of model are described as:

- **Plain Networks:** The designed plain networks are based on VGG Network for image recognition. The layers having convolutions mostly have 3x3 size and for same number of size same output feature map size, the layers have the same number of filters. Number of filters are doubled, when the map size is halved. This is because to keep the lower complexity per layer. By stride of 4, we perform down sampling so as to reduce layer data size and complexity. No. of weighted layers in the image is 34, and is expandable as in v1 for 128, v2 for 256, v3 for 512, and v4 for 1024.

- **Block Networks:** Based upon plain networks, but each and every layer is blocked by a bunch of predetermined set of layers; called as blocks. Each block has an input size 256 x 256 of data image and outputs a similar size image with convoluted data weights and function approximations. Each block has modified batch normalization (BN) layer, a loss function layer (AM softmax variant with linear weights), 3x3 convolutional layers, and stochastic pooling layer.

## 4 Results

### 4.1 Accuracy Metric

We use Top 1 accuracy as a comparison metric between performance of different models and architectures. Top 1 accuracy is the accuracy when a model gives a single accuracy probability and is match with the trained dataset accuracy. There are other variants of accuracy metric such as Top 5 accuracy and Top 10 accuracy but they give 5 and 10 values of probabilities respectively and out of which the closest one is to be compared; the disadvantage of Top 5 and Top 10 accuracy is that they might produce false results since the accuracy range is pretty narrow in ImageNet based benchmark we created, and as from ImageNet competition.

Since we created 4 models with Alpha-Net architecture: v1, v2, v3, and v4 with different layers, we analyze all the possible combinations arrive over different parameter and technique.

### 4.2 Comparison with Layer Structure

The comparison of all the Alpha-Net models with respect to layer structure is shown in Table 6.1 with layer structures as plain layer structure, residual block structure, and alpha block structure. We used Alpha blocks for our benchmark.

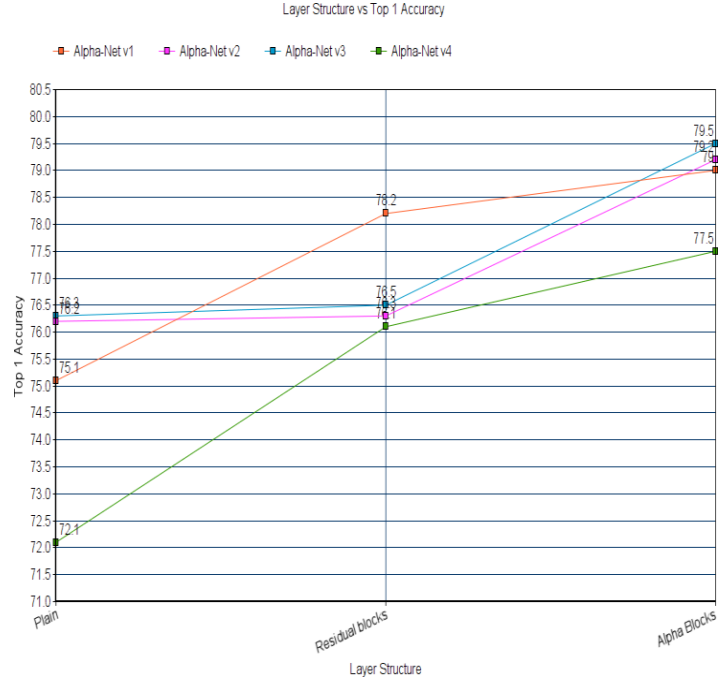

Figure 7: Graph showing layer structure vs Top 1 accuracy (%). Accuracy values of Alpha-Net v1, v2, and v3 is greater than v4 because of overfitting. This proves that for different layer structures for common architecture, increasing the number of layers does not increases training accuracy linearly, at some point training accuracy decreases.

| Architecture | No. of Layers | Layer structure (Top 1 Accuracy) | | |
|---|---|---|---|---|
| | | Plain | Residual Blocks | Alpha Blocks |
| Alpha-Net v1 | 128 | 75.1% | 78.2% | 79.0% |
| Alpha-Net v2 | 256 | 76.2% | 76.3% | 79.2% |
| Alpha-Net v3 | 512 | 76.3% | 76.5% | 79.5% |
| Alpha-Net v4 | 1024 | 72.1% | 76.1% | 77.5% |

Table 1: Accuracy comparison of Alpha-Net models vs Layer structure.

Best results are shown by alpha-blocks because of careful formulation of layers inside the alpha-block. We can also see that the accuracy of v4 model is less than its expected accuracy because of the possible reason of overfitting with large number of layers in them.

### 4.3 Comparison with Loss function

The comparison of all the Alpha-Net models with respect to loss functions is shown in Table 6.2 with loss functions as Softmax function, AM Softmax function, AM Softmax function with linear weights. We used AM Softmax function with linear weights for our benchmark.

Best results are shown by AM Softmax function with linear weights because of linear weights improvements over normal Additive margins. We can also see that the accuracy of v4 model is less than its expected accuracy because of the possible reason of overfitting with large number of layers in them.

### 4.4 Comparison with Normalization function

The comparison of all the Alpha-Net models with respect to the normalization function is shown in Table 6.3 with normalization functions as log scaling, z-score, and Alpha-encoding. We used Alpha encoding for our benchmark.

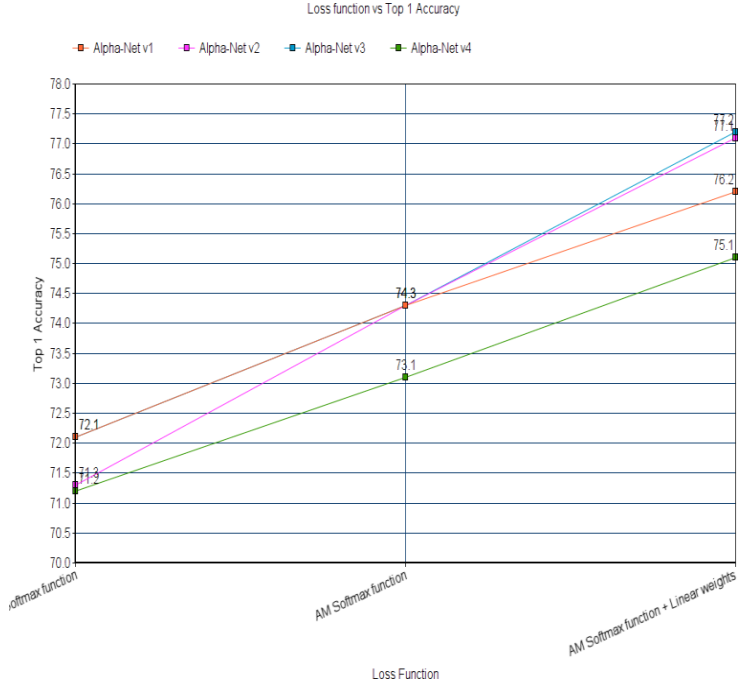

Figure 8: Graph showing loss function vs Top 1 accuracy (%). Alike with layer structure, accuracy values of Alpha-Net v1, v2, and v3 is greater than v4 because of overfitting. This proves that for different layer structures for common architecture, increasing the number of layers does not increases training accuracy linearly, at some point training accuracy decreases.

| Architecture | No. of Layers | Softmax | Loss Function (Top 1 Accuracy) | |
|---|---|---|---|---|
| | | | AM Softmax | AM Softmax + Linear Weights |
| Alpha-Net v1 | 128 | 72.1% | 74.3% | 76.2% |
| Alpha-Net v2 | 256 | 71.3% | 74.3% | 77.1% |
| Alpha-Net v3 | 512 | 72.1% | 74.3% | 77.2% |
| Alpha-Net v4 | 1024 | 71.2% | 73.1% | 75.1% |

Table 2: Accuracy comparison of Alpha-Net models vs Loss function.

Best results are shown by Alpha encoding because of simple and consistent feature extraction by alpha-transformations. We can also see that the accuracy of v4 model is less than its expected accuracy because of the possible reason of overfitting with large number of layers in them.

### 4.5 Comparison between state-of-the-art architectures

The comparison of all the Alpha-Net models (v1, v2, v3, and v4) with different architectures is shown in Table 6.4. The global benchmark is pop out to be InceptionResNet v2. InceptionResNet v2 is the second improvement with combinational features of residual blocks and representation with Inception module functionalities.

| Architecture | No. of Layers | Normalization (Top 1 Accuracy) | | |
|---|---|---|---|---|
| | | log-scaling | z-score | Alpha-encoding |
| Alpha-Net v1 | 128 | 69.2% | 71.2% | 71.0% |
| Alpha-Net v2 | 256 | 69.5% | 70.1% | 71.2% |
| Alpha-Net v3 | 512 | 70.1% | 70.1% | 71.5% |
| Alpha-Net v4 | 1024 | 71.2% | 69.5% | 70.5% |

Table 3: Accuracy comparison of Alpha-Net models vs Normalization function.

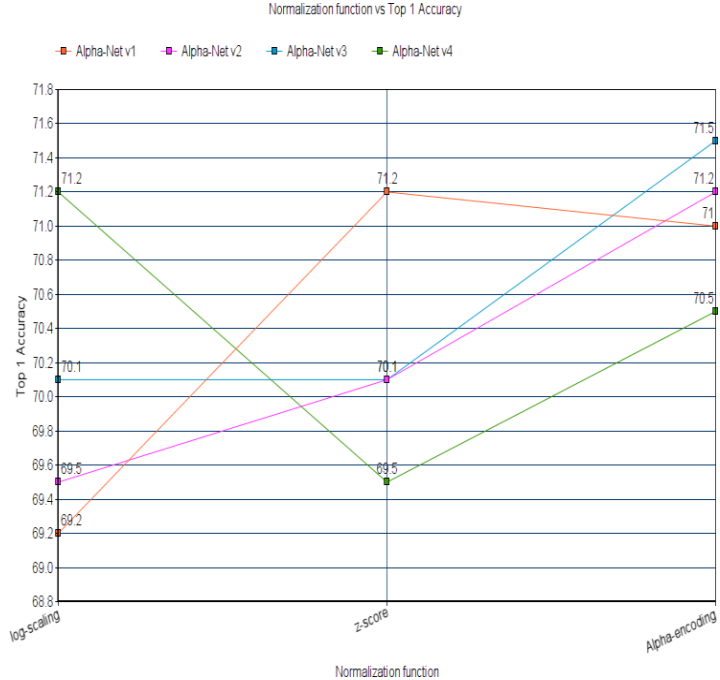

Figure 9: Graph showing normalization function vs Top 1 accuracy (%). Accuracy values of Alpha-Net v1, v2, and v3 is greater than v4 because of overfitting. This proves that for different layer structures for common architecture, increasing the number of layers does not increases training accuracy linearly, at some point training accuracy decreases.

| Architecture | Top 1 Accuracy |
| --- | --- |
| Xception | 79.0% |
| Inception v3 | 78.8% |
| ResNet 50 | 75.9% |
| VGG 19 | 72.7 |
| VGG 16 | 71.5 |
| InceptionResNet v2 | 80.4% |
| Alpha-Net v1 | 78.2% |
| Alpha-Net v2 | 79.1% |
| Alpha-Net v3 | 79.5% |
| Alpha-Net v4 | 78.3% |

Table 4: Accuracy comparison of various architectures over ImageNet based benchmark.

Our benchmark produces **second best results for v3**, and are accompanied by AM Softmax function with linear weights, Alpha-encoding, and Alpha-blocks in combination.

## 5 Conclusion

Deep learning systems are complex and non-comprehensive by their nature of deep layers and data representations. Combining powerful features to a single layer and processing is a computationally expensive task. Much of which require a cloud platform (GCP or AWS) for processing and storage. A new architecture is proposed named Alpha-Net based on the alpha-transformations of the input data; which is important because to keep data size low for faster processing. Four custom models were implemented based upon Alpha-Net named Alpha-Net v1, v2, v3, and v4 with layers 128, 256, 512, and 1024. The common misconception that increasing layer size increases the training accuracy is also successfully busted when v4 model shows less accuracy than v1, v2, and v3; that means the threshold number of layers is in between v3 and v4; i.e. 512 and 1024. Since, some authors have proved that ResNet has some redundant layers; we used this fact to improve the mapping scheme of the layers based on stochastic approach. The data flow is not linear, and certainly not many to many (increases the training complexity), so we randomly assigned mapping from one layer to all other layers, which adjusts itself in many iterations.

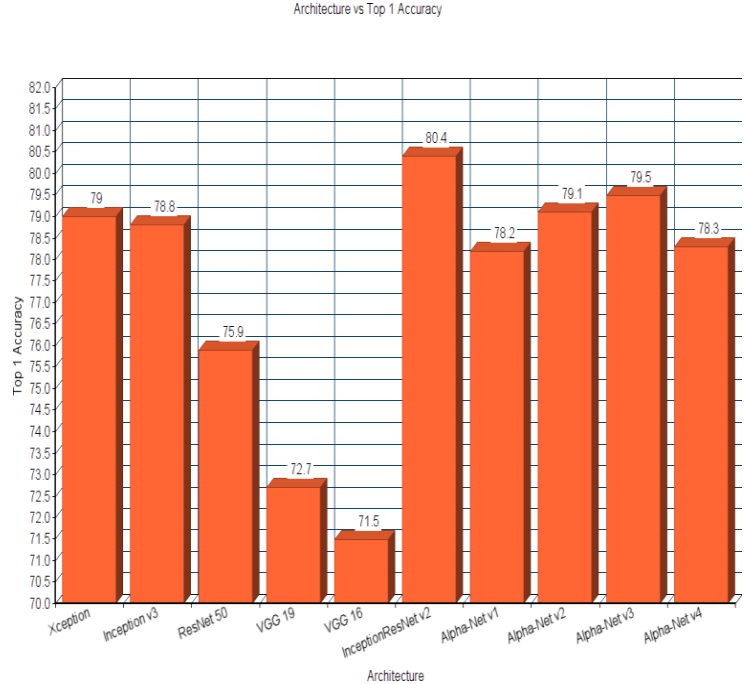

Figure 10: Graph showing various architectures vs Top 1 accuracy (%). The maximum accuracy is shown by InceptionResNet v2 of 80.4% and among Alpha-Net models best accuracy is shown by v3 of 79.5%; which is second best Top 1 accuracy shown by analyzed state-of-the-art models of ResNet and Inception architecture.

The project has many **novel contributions** including but not limited to the dataset used, input representation, normalization of input data, layer mappings, block structure of layers, and the loss function – linear weights + Additive Margin Softmax function. The result of each and every novelty affects the training accuracy in an indifferent manner; the summary of which is described in results analysis.

**From our results analysis, Alpha-Net performs better than ResNet (original and its variants) on ImageNet based benchmark.** Multiple trainings of alpha-net models suggest that the optimal number of layers for any computational training task is constant, and is often less than 10,000.

# 6  Acknowledgement

This work was done as the major project for Prof. Bholanath Roy and Prof. Sweta Jain at Maulana Azad National Institute of Technology, Bhopal. The authors would like to thank Bholanath Roy, Sweta Jain, Saritha Khetawat, Deepak Singh Tomar, and Sanyam Shukla for their valuable suggestions regarding the work. Authors are also grateful to their fellow mates for healthy competition and criticism.

## Appendix - A: Tools and Technologies Used

**Minimum System/Software/Hardware requirements**

- Windows 10 / Linux / Mac OS / Chromium OS.

- Google Cloud Platform (GCP) for faster trainings and processings (If local system not compatible)
- Memory requirements: 16 GB (RAM) with a GPU or TPU.
  - Recommended NVIDIA GEFORCE 940+MX
- Secondary memory requirements (Hard disk): 200 GB (ROM).
  - Recommended 512 GB / 1 TB for dataset storage and processing
  - HDFS if distributed is necessary
- Latest versions of web browsers - Chrome, Chromium, Firefox, or Opera.
- Python libraries: Scikit-learn, Numpy, Pandas, Python 3.8
- Writer and diagram designing software such as LibreOffice Draw.
- Clock Speed: 1.5 Ghz
- Virtual Memory: 32 bits (minimum).
- Cache Memory: 512 MB, etc.

**Resources Usage**
- Operating System: Windows 10 / Ubuntu 20.04 / **Kali Linux 20.0.4**
- Memory: **8 GB** (RAM)
- Secondary memory: **1 TB** (ROM)
- Firefox, Chrome, **Chromium**, and Opera web browsers.
- Microsoft Word 16 / **LibreOffice Writer 5.4** (Linux).

**Functional Requirements**
- Cached data storage and retrieval.
- Proper User Interface as in accordance with common use-cases; i.e. the terminal.
- Customized testing functionalities in accordance with the proposed model (Python3.8).

## Appendix - B: Implementation and Coding Details

**Coding details**

Our implementation for the proposed dataset follows the standard pratice of training and testing methodology based on Keras v2. The image is resized with its shorter side randomly sampled in [256, 480] for scale augmentation. All the images with faces are aligned so as to minimize the complexity of the angle and orientation of the image. Augmentation of color is used as described in original ResNet paper. We take convolutional layer, activation layer and batch normalization to form a block, in our alpha blocks as similar to residual blocks. We initialize the weights as in the ImageNet dataset and train the network. The learning rate ($\beta$) starts from 0.01 and is divided by 10 when the error converges, and the models are trained for up to 60x10, 8 for each v1, v2, v3, and v4. For consistency reasons, dropout was not used.

Parameters used for training are as – Momentum: 0.9, Batch size division: 10, Weight decay: 0.0001, Initial learning rate: 0.01, Batch size: 512

**Testing**

For testing purposes, we used 10 crop technique for qualitative analysis. The data is first normalized as per the image size and is used as per our transformations, and the scores are averaged over certain image sizes such as 32, 64, 128, 256, and 512. It is also important to note that stochastic pooling for custom objects is not available in Keras, so we implemented it separately for convenience.